\pdfoutput=1
\documentclass[10pt,twocolumn,letterpaper]{article}

\usepackage{wacv}
\usepackage{times}
\usepackage{epsfig}
\usepackage{graphicx}
\usepackage{amsmath}
\usepackage{amssymb}
\usepackage{booktabs}
\makeatletter
\@namedef{ver@everyshi.sty}{}
\makeatother
\usepackage{tikz}
\usepackage{comment}
\usepackage{color}
\usepackage{subcaption}
\usepackage{caption}
\usepackage{booktabs}
\usepackage{multirow}
\usepackage{wrapfig}

\def\wacvPaperID{286} 

\wacvfinalcopy 

\usepackage[pagebackref=true,breaklinks=true,colorlinks,bookmarks=false]{hyperref}

\usepackage[capitalize]{cleveref}
\crefname{section}{Sec.}{Secs.}
\Crefname{section}{Section}{Sections}
\Crefname{table}{Table}{Tables}
\crefname{table}{Tab.}{Tabs.}

\hypersetup{
  citecolor=blue,
}

\newcommand\nnfootnote[1]{%
  \begin{NoHyper}
  \renewcommand\thefootnote{}\footnote{#1}%
  \addtocounter{footnote}{-1}%
  \end{NoHyper}
}

\newcommand{\re}[1]{\textcolor{red}{#1}}
\newcommand{\bl}[1]{\textcolor{blue}{#1}}

\begin{document}

\title{Keys to Better Image Inpainting: Structure and Texture Go Hand in Hand}

\author{Jitesh Jain\textsuperscript{1,2,3*$\dag$} \quad Yuqian Zhou\textsuperscript{4*$\dag$} \quad Ning Yu\textsuperscript{5} \quad Humphrey Shi\textsuperscript{1,3} \\ \\ 
{\textsuperscript{1}SHI Lab @ University of Oregon \qquad \textsuperscript{2}IIT Roorkee} \\ \textsuperscript{3}Picsart AI Research (PAIR) \qquad \textsuperscript{4}Adobe Inc. \qquad \textsuperscript{5} Salesforce Research \\
{\small \url{https://github.com/SHI-Labs/FcF-Inpainting/}}
}

\twocolumn[{%
\renewcommand\twocolumn[1][]{#1}%
\maketitle
\begin{center}
    \centering
    \captionsetup{type=figure}
    \includegraphics[width=\textwidth]{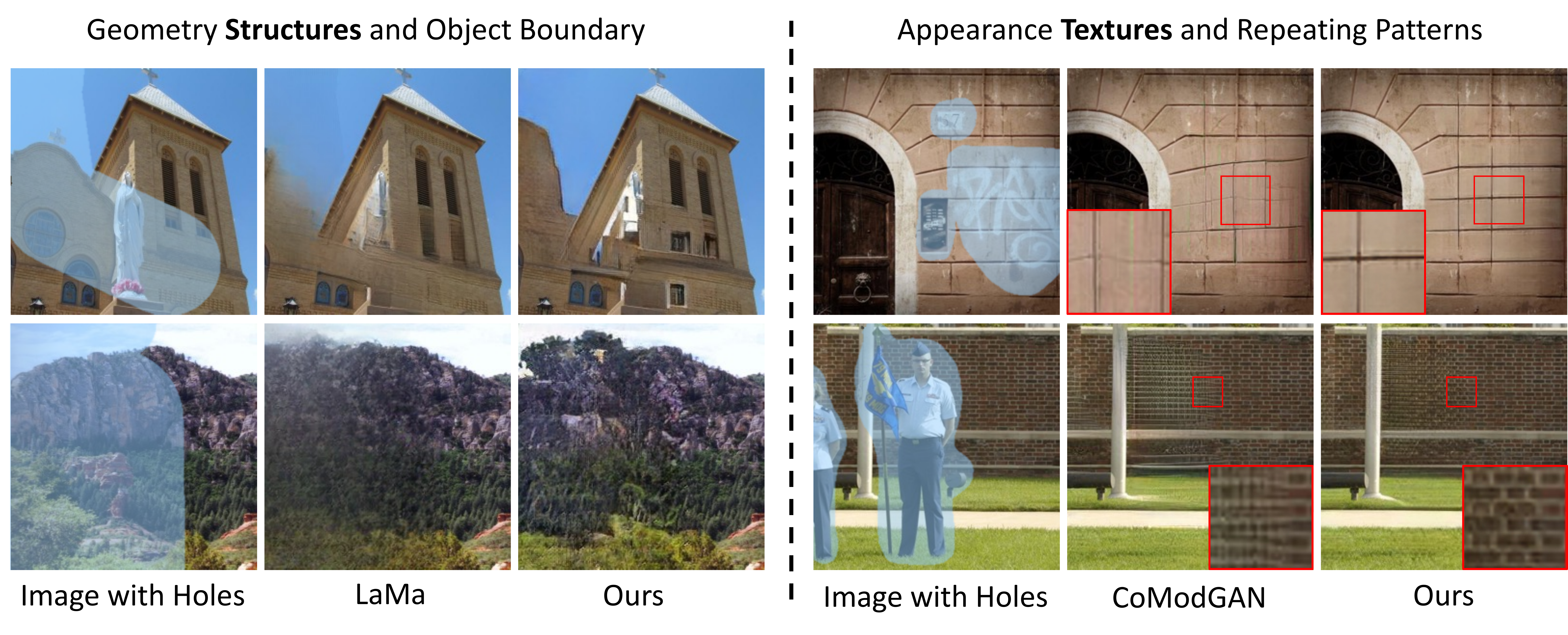}
    \captionof{figure}{The most challenging issues for advanced image inpainting algorithms fall on generating better \textbf{structures} and \textbf{textures}. Left: LaMa \cite{lama} works well for repeating textures but generates fading out boundaries and structures when the holes get larger. Right: CoModGAN ~\cite{comodgan} with a StyleGAN-based \cite{stylegan2} generator achieves impressive geometry structures, but it fails to reuse textures within the image to generate plausible repeating patterns. Our model generates good structures and textures simultaneously better than any state-of-the-arts.}
    \label{fig:demo}
\end{center}%
}]

\nnfootnote{$^*$ Equal Contribution.}
\nnfootnote{$^\dag$ This work started when Jitesh interned at SHI Lab @ University of Oregon, and Yuqian was a Ph.D. student at IFP @ UIUC.}

\begin{abstract}

Deep image inpainting has made impressive progress with recent advances in image generation and processing algorithms. We claim that the performance of inpainting algorithms can be better judged by the generated structures and textures. Structures refer to the generated object boundary or novel geometric structures within the hole, while texture refers to high-frequency details, especially man-made repeating patterns filled inside the structural regions. We believe that better structures are usually obtained from a coarse-to-fine GAN-based generator network while repeating patterns nowadays can be better modeled using state-of-the-art high-frequency fast fourier convolutional layers. In this paper, we propose a novel inpainting network combining the advantages of the two designs. Therefore, our model achieves a remarkable visual quality to match state-of-the-art performance in both structure generation and repeating texture synthesis using a single network. Extensive experiments demonstrate the effectiveness of the method, and our conclusions further highlight the two critical factors of image inpainting quality, structures, and textures, as the future design directions of inpainting networks.




\end{abstract}
\section{Introduction}

Image inpainting aims to fill in missing parts of the incomplete input image such that an observer cannot distinguish between the inpainted regions and real regions of the output image. It has many applications in the industry, like object removal, photo retouching, and old photo restoration.

Traditionally, inpainting is achieved by diffusion-based \cite{bertalmio2000image} or patch-based methods \cite{patchmatch}. They assume that the missing contents inside the hole regions can be synthesized by reusing textures or colors from the same image. These methods, especially patch-based ones, synthesize remarkable textures but mostly fail to complete semantic structures within the hole. GAN-based methods \cite{deepfillv1,deepfillv2} make the semantic structure generation possible. Among them, DeepFill \cite{deepfillv1,deepfillv2} first considered structure and texture synthesis. The model follows a two-stage network, where the first stage generates a coarse semantic map, and the second stage utilizes global contextual attention to copy similar deep features for texture enhancement. However, for most previous deep inpainting models \cite{lahiri2020prior,yi2020contextual,zhao2020uctgan,zhou2020learning,wang2020vcnet,liao2020guidance,iccv_ctsdg,aotgan,yuqian2022image,transfill,geofill}, when the hole gets larger, the structure estimation becomes challenging. Recently, two milestones, LaMa \cite{lama}, and CoModGAN \cite{comodgan} inspire us to study further the capability of deep networks handling inpainting structures and textures.

Zhao \emph{et al.} \cite{comodgan} proposed the Co-Modulated Generative Adversarial Network (CoModGAN), which augments the encoded representation with a mapped stochastic noise vector to allow for stochasticity and better generation quality for images with large holes. The impressive image generation capability sources from the StyleGAN2-based \cite{stylegan,stylegan2} generator, which follows a coarse-to-fine scheme. The conditional StyleGAN2 feeds the incomplete image's global style and coarse structures and leverages rich structure data in the training dataset for novel generations. However, the generation quality relies more on the training data domain. Since CoModGAN does not include attention-related structures to enlarge the receptive field, the original image textures cannot be fully reused. CoModGAN performs poorly with novel textures and man-made repeating patterns.

The trend of image inpainting has been expected to be changed since LaMa~\cite{lama} came out. Suvorov \emph{et al.} \cite{lama} used the \textit{Fast Fourier convolution}~\cite{ffc} inside their ResNet based LaMa-Fourier model to account for the lack of receptive field for generating repeating patterns in the hole regions. Before that, researchers struggled with global self-attention \cite{deepfillv2} and its high computational cost, but still cannot achieve reasonable recovery for repeating man-made structures as good as LaMa. Nevertheless, LaMa generates fading-out structures when the hole becomes larger and across object boundary. Recently, transformer-based methods \cite{zheng2021tfill,wan2021high} model the global attention, while the structures can only be computed within a low-resolution coarse image. Beyond that, good repeating textures cannot be synthesized. Recent diffusion-based inpainting models \cite{ldm, palette, repaint} pushed the limits of generative models, but the inference time can be too long for practical usage. 

In this paper, we revisited the core design ideas of state-of-the-art deep inpainting networks. 
To address the issues mentioned above, we propose an intuitive and effective inpainting architecture that augments the powerful co-modulated StyleGAN2 \cite{stylegan2} generator with the high receptiveness ability of FFC \cite{ffc} to achieve equally good performance on both textures and structures as shown in \cref{fig:demo}. Specifically, we generate image structures in a coarse-to-fine StyleGAN-based generation scheme. Meanwhile, we merge between the generated coarse features and the skip features from the encoder and pass them through a Fast Fourier Synthesis (FaF-Syn) module to better generate repeating textures. Our idea is simple yet effective, making structures and textures well synthesized within a single network.  

To summarize, we find that better structures can be obtained in a deeper coarse-to-fine GAN-based generator, and repeating textures are better synthesized with multi-scale high receptive field fourier convolutional layers. We combine the advantages of the two and propose a Fourier Coarse-to-Fine (FcF) generator for general-purpose image inpainting. Our model well handles textures and structures simultaneously and generalizes well to both natural and man-made scenes. Extensive experiments demonstrate the effectiveness of our proposed framework, and it achieves a new state-of-the-art on the CelebA-HQ dataset and remarkable performance comparable to state-of-the-arts on the Places2 dataset with a higher user preference rate.

\section{Related Work}

\noindent\textbf{Traditional Image Inpainting.} 
Traditionally, image inpainting tasks were resolved by either diffusion-based or exemplar-based methods. Diffusion-based methods use PDEs \cite{bertalmio2000image, a1, a2} or variational methods \cite{ballester2001filling, a3} etc. to fill the hole by propagating the image pixels from the non-hole regions into the missing regions. It usually works well for connecting lines, curves within thinner holes since the smoothness constraints are used to regularize the hole-filling process, while for larger holes with more ambiguous structures, those approaches easily generate blur results. Copy-pasting of similar patches within the same images is categorized as the exemplar-based synthesis method. Both pixel-wise copying \cite{a4, a5} and patch-based synthesis \cite{patchmatch,xu2010image, a6} suffer from expensive nearest neighbor searching. Those methods produce good textures but messy structures.

\begin{figure}[t]
\centering
\includegraphics[width=1\linewidth]{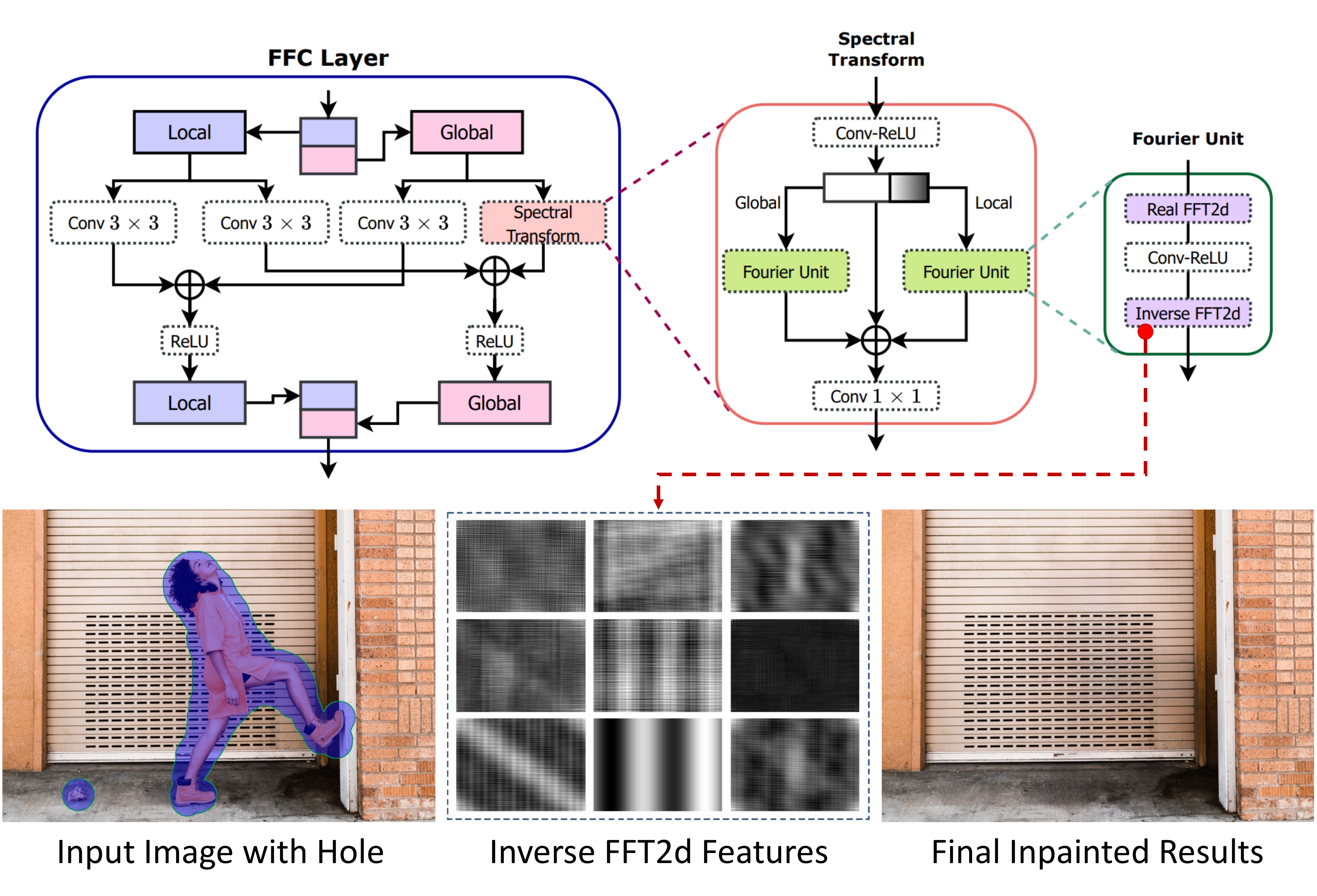}
\vspace{-0.8cm}
\caption{\textbf{FFC Layer and InverFFT2d feature visualization.} The FFC uses the Spectral Transform module in the global branch to account for the global context similar to LaMa~\cite{lama}. The learned inverse FFT2d layer features explain why LaMa works well on repeating patterns. It actually generates global repeating patterns but not reconstructs image contents. The learned global repeating patterns are further merged inside the hole region to synthesize more complicated repeating patterns. }
\label{fig:ffc_inv}
\vspace{-0.5cm}
\end{figure}

\noindent\textbf{Deep Image Inpainting.} 
GAN-based \cite{gan} deep generative models \cite{lahiri2020prior,yi2020contextual,zhao2020uctgan,zhou2020learning,wang2020vcnet,liao2020guidance,iccv_ctsdg,aotgan} were widely applied to inpainting tasks recently. Pathak \emph{et al.} \cite{pathak2016context} first attempted to use GAN to address filling holes using semantically consistent contents. EdgeConnect~\cite{edge-connect} proposed to use edge detection results as the guidance for inpainting to form better structures. Later on, partial convolution ~\cite{partialconv} and gated convolution \cite{deepfillv1, deepfillv2} are proposed to tailor deep generative model for incomplete image feature extraction and reusing, making deep inpainting work for free-form holes. ProFill \cite{profill} then extended deepfillv2 \cite{deepfillv2} to apply iterative filling and confidence estimation to refine the textures. These methods fail to perform well on large irregular masks and textural images due to the small receptive field of the generator, lack of stochasticity, or larger memory and speed concerns like contextual attention \cite{deepfillv1}. Following the success of GAN-based methods, we formulate our inpainting framework based on the StyleGAN2~\cite{stylegan2} architecture. The image generation ability associated with StyleGAN2 from stochasticity enables the large-hole filling with realistic structures.

\noindent\textbf{Large Mask Inpainting.} Recently,  CoModGAN~\cite{comodgan} proposed a co-modulation strategy using stochastic noise inside conditional StyleGAN2~\cite{stylegan2} for improving image generation ability for large hole inpainting. Still, CoModGAN does not perform well when tested on texture-based images. For tackling repeating patterns in images, LaMa~\cite{lama} proposed the use of Fast-Fourier Convolutions~\cite{ffc} inside the generator structure. However, LaMa produces a smooth and faded effect for large continuous masks. More recently, CMGAN~\cite{zheng2022cmgan} used FFC inside encoder and a cascaded global-spatial modulation-based decoder along with training on object-aware masks. However, CMGAN~\cite{zheng2022cmgan} struggles at good structure generation.  In this work, we propose to combine the benefits of FFC and stochasticity using noise inside the co-modulated StyleGAN2~\cite{stylegan2} coarse-to-fine generator to achieve robust performance on both textural and structural images for large free-form masks. The unification of stochasticity in co-modulated StyleGAN2 and FFC is non-trivial. It requires careful design of an architecture that does not collapse and behave effectively. Extensive experiments demonstrate that our integration prevents FFC from magnifying the noise in the coarse-level layers. 



\section{Methodology}
\label{sec:method}

\begin{figure*}[t!]
\centering
\includegraphics[width=0.9\linewidth]{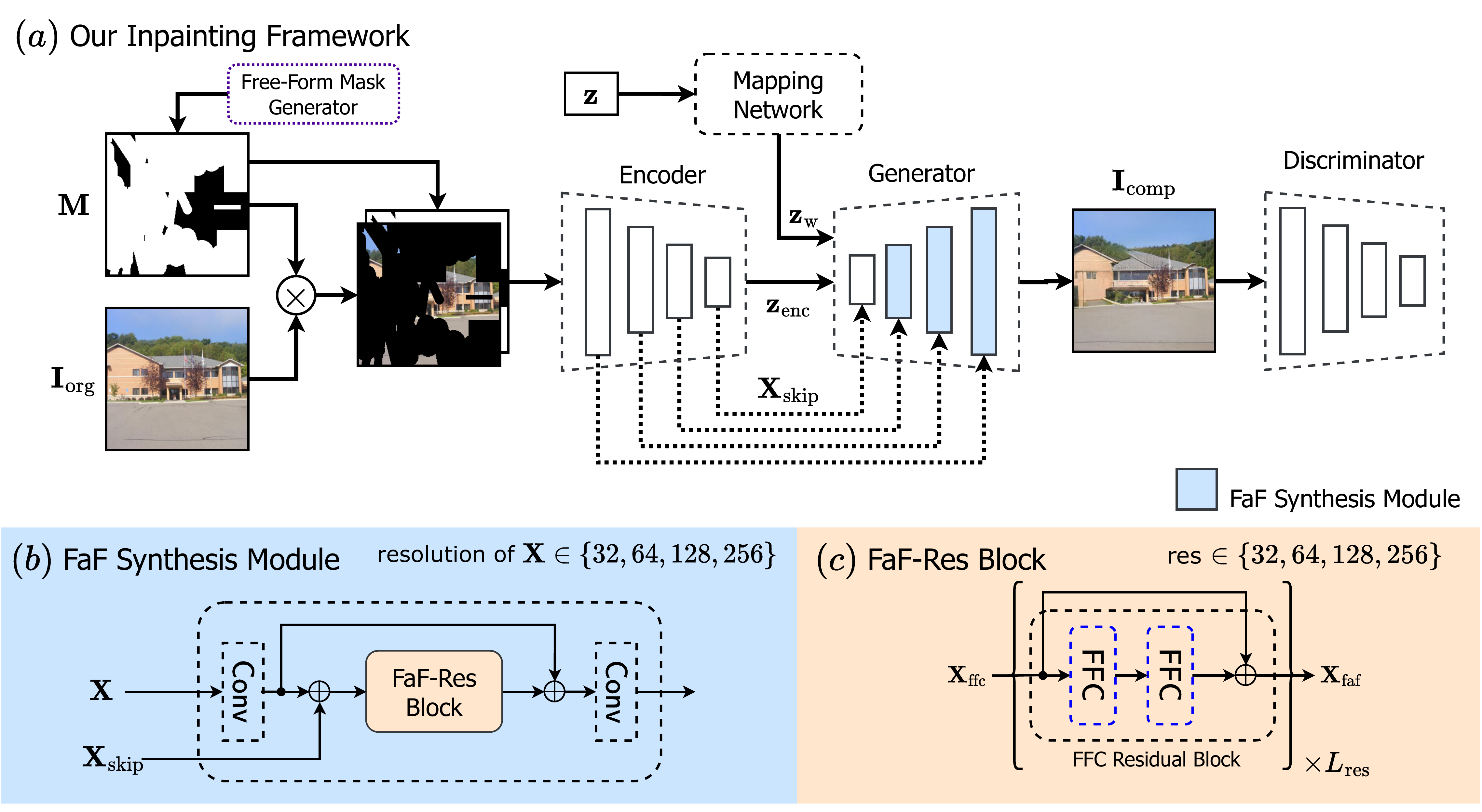}
\vspace{-2mm}
\caption{\textbf{Our model architecture.} (a) The inpainting framework. (b) The architecture of our FaF Synthesis (FaF-Syn) module inside the generator for resolutions $\in [32,64,128,256]$. The convolutional layers inside FaF-Syn are co-modulated using the encoded features and style mapping of the latent noise vector. (c) The architecture of our FaF-Res Block.}
\label{fig:ffc_gan}
\vspace{-0.5cm}
\end{figure*}

In this section, we introduce the newly proposed network architecture, as shown in \cref{fig:ffc_gan}. The four-channel inputs concatnate the RGB masked image ($\mathbf{I}_\text{hole}$) and the hole  ($\mathbf{M}$), where $\mathbf{I}_\text{hole} = \mathbf{I}_\text{org} \odot (1 - \mathbf{M})$. The inputs are fed into the encoder network ($\mathcal{E}$) to obtain the encoded latent vector $\mathbf{z}_\text{enc}$ and multi-level feature maps $\mathbf{X}_\text{skip}$. Our generator network ($\mathcal{G}$) shares the spirit of the StyleGAN2 \cite{stylegan2} architecture. Similar to CoModGAN~\cite{comodgan}, we generate random noise latent vector $\mathbf{z}$ and pass it through a mapping network ($\mathcal{M}$) to obtain the embedding $\mathbf{z}_\text{w}$. $\mathbf{z}_\text{w}$ is concatenated with $\mathbf{z}_\text{enc}$ and fed into the generator $\mathcal{G}$. The core contribution is that we newly propose the Fast Fourier Synthesis Module (FaF-Syn) inside the Fourier Coarse-to-Fine (FcF) generator. More intuitions and details are introduced below. 
\subsection{Fourier Coarse-to-Fine (FcF) Generator}
We aim to integrate the idea of LaMa, fast fourier convolutional residual blocks, into a co-modulated StyleGAN2-based coarse-to-fine generator. Intuitively, the coarse-to-fine generator renders global structures and image styles from the high-level feature and noise embedding. During the upsampling process in the generator, global texture features, both in the non-hole regions and in the generated hole regions, can be extracted by fast fourier convolutional layers and integrated appropriately to refine textures within the randomly generated structures. The idea is realized by a Fast Fourier Synthesis (FaF-Syn) module consisting of a Fast Fourier Residual (FaF-Res) Block. Within each FaF-Res block, there are two Fast Fourier Convolutional (FFC) layers. We will introduce them in a bottom-up order. \\
\noindent\textbf{Fast Fourier Convolutional Residual Blocks (FaF-Res).}
The FaF Residual block in \cref{fig:ffc_gan} (c) consists of two Fast Fourier Convolutional (FFC) layers (\cref{fig:ffc_inv}). The FFC \cite{ffc} layer is based on a channel-wise fast Fourier transform (FFT) \cite{fft}. It splits channels into two branches: a) local branch uses conventional convolutions to capture the spatial details, and b) global branch uses a Spectral Transform module to consider the global structure and capture the long-range context. Finally, the outputs of the local and global branches are stacked together.

The Spectral Transform uses two Fourier Units (FU) to capture the global and semi-global information. The left Fourier Unit (FU) models the global context. On the other hand, the Local Fourier Unit (LFU) on the right side takes in a quarter of the channels and focuses on the semi-global information in the image. A Fourier Unit mainly breaks down the spatial structure into image frequencies using a Real FFT2D operation, a convolution operation in the frequency domain and finally recovering the structure using an Inverse FFT2D operation.

LaMa first applied FFC layers in inpainting yet did not reveal the reasons why it works for successfully synthesizing repeating patterns. We analyze LaMa's intermediate features within the FFC layers and find that, after the inverse FFT2D layer within the Fourier Unit, the learned features do not represent and reconstruct complicated image contents directly, but generating multiple global repeating patterns, as shown in \cref{fig:ffc_inv}. The learned global repeating patterns are then merged inside the hole region to synthesize more complicated repeating contents. Therefore, in order to use FFC more effectively for inpainting, it is better to integrate the FFC layers into the generation process instead of feature encoding. It inspires us to carefully design a multi-scale FFC synthesis block and incorporate the FFC layers into the coarse-to-fine generator parts of the StyleGAN2.\\
\noindent\textbf{Fast Fourier Synthesis (FaF-Syn) Module.}
Our generator ($\mathcal{G}$) shares a similar idea with CoModGAN~\cite{comodgan}, but the main difference is that we design the newly proposed Fast Fourier Synthesis (FaF-Syn) module (\cref{fig:ffc_gan}(b)) inside the coarse-to-fine generation process. 

Integrating it into a StyleGAN2-based generator is not trivial. There are two main problems to consider: First, the global repeating textures can be better modeled from the encoding features or in the generated features via a skip connection. Should we embed the FFC blocks into the encoder or the generator? We assume that it's better to utilize it in the generation process by visualizing and analyzing the FFC features. Second, suppose we integrate the FFC blocks into the generator, the FFC layers may magnify the noisy generated structures in the very coarse level layers, causing unstable training and harming the performance. Which level of features will be better to include the FFC layers?

We empirically formulate our network in the following way: First, we use skip connections between $\mathcal{E}$ and $\mathcal{G}$ layers corresponding to the same resolution scale. 
Second, we introduce this Fast Fourier Synthesis (FaF-Syn) module as shown in Figure \ref{fig:ffc_gan}. FaF-Syn takes in both the encoded skip connected features $\mathbf{X}_\text{skip}$, and the features $\mathbf{X}_\text{skip}$ upsampled from the previous level in the generator. FaF-Syn explicitly integrates the features from the encoder (\textit{i.e.} existing image textures) and the generator (\textit{i.e.} generated textures from the previous layers) to synthesize the global repeating textural features. It allows us to take advantage of the previous coarse-level repetitive textures and further refine them at the finer level. FaF-Syn is only applied to feature resolutions of $32\!\times\!32$, $64\!\times\!64$, $128\!\times\!128$, and $256\!\times\!256$. Our experiments show that applying it to a coarse level (like $8\!\times\!8$ and $16\!\times\!16$) harms the performance (Supplementary Material). 

\subsection{Other Modules}
\noindent
\textbf{Encoder Network.}
Our encoder ($\mathcal{E}$) follows a similar architecture to the discriminator used in StyleGAN2 \cite{stylegan2} but without the residual skip connections. $\mathcal{E}$ takes $\mathbf{I}_\text{hole}$ and $\mathbf{M}$ downsamples it to a spatial size of $4\!\times\!4$. We also use skip connections between $\mathcal{E}$ and $\mathcal{G}$. Finally, we pass the flattened $4\!\times\!4$ encoded feature map through a linear layer to obtain an encoded latent vector $\mathbf{z}_\text{enc}$. 


\noindent\textbf{Mapping Network.}
We use a mapping network ($\mathcal{M}$) in our framework to transform our noise latent vector ($\mathbf{z} \sim \mathcal{N}(\mathbf{0},\mathbf{I})$) to an a latent space $\mathbf{z}_\text{w} = \mathcal{M}(\mathbf{z})$ \cite{stylegan2}. We further perform an affine transform ($\mathcal{A}$) on a concatenation of $\mathbf{z}_\text{w}$ and $\mathbf{z}_\text{enc}$ from $\mathcal{E}$ as $\mathbf{s} = \mathcal{A}\big(\texttt{stack}(\mathbf{z}_\text{enc}, \mathbf{z}_\text{w})\big)$. The style coefficient ($\mathbf{s}$) from $\mathcal{A}$ is used to scale the weights of the convolutional layers inside our generator ($\mathcal{G}$). The architecture of $\mathcal{M}$ is similar to the 8-layer MLP mapping network used in StyleGAN2~\cite{stylegan2}.

\noindent \textbf{Discriminator.}
For our discriminator, we stick to the residual discriminator proposed in StyleGAN2~\cite{stylegan2}. Our discriminator takes in a concatenation of hole masks $\mathbf{M}$ and the original image $\mathbf{I}_\text{org}$ or the completed image $\mathbf{I}_\text{comp}$ depending on the training phase.


\subsection{Loss Functions}
We utilize a non-saturating logistic loss \cite{gan} with an R1-regularization \cite{r1_loss} for our adversarial losses. We also use a reconstruction loss along with a high receptive field perceptual loss \cite{lama} to supervise the structures in the images during training. We find that reconstruction loss is important for learning the repeating patterns using FFC and the proposed FaF-Syn module.\\
\noindent
\textbf{Adversarial Loss.}
In the similar spirit of \cite{stylegan2}, we use the non-saturating cross-entropy loss for the adversarial training of our inpainting framework. The input of the discriminator is the concatenation of $\mathbf{M}$ and real $\mathbf{I}_\text{org}$ or fake $\mathbf{I}_\text{comp}$. 

\noindent
\textbf{High Receptive Field Perceptual Loss.}
For the loss of the generator, similar to LaMa, we use a high receptive field perceptual loss (HRFPL) \cite{lama} which computes the $\ell_2$ distance between $\mathbf{I}_\text{comp}$ and $\mathbf{I}_\text{org}$, after mapping these images onto higher level features. The feature extractor is based on dilated ResNet-50~\cite{multidilated,yu2017dilresnet} and is pretrained for ADE20K~\cite{zhou2017scene,zhou2018semantic} semantic segmentation. Similar to \cite{partialconv}, the loss can be represented as $\mathcal{L}_\text{HRFPL} = \sum_{p=0}^{P-1} {\frac{\|{\Psi}_{p}^{\mathbf{I}_\text{comp}}-{\Psi}_{p}^{\mathbf{I}_\text{org}}\|_{2}}{N}}$, where 
${\Psi}_{p}^{\mathbf{I}_\text{*}}$ is the feature map of the $p^{th}$ layer given an input $\mathbf{I}_\text{*}$, where $N$ is the number of feature points in ${\Psi}_{p}^{\mathbf{I}_\text{org}}$. 

\noindent
\textbf{Total Loss.}
We also include a pixel-wise reconstruction $\ell_1$ loss between $\mathbf{I}_\text{comp}$ and $\mathbf{I}_\text{org}$: $\mathcal{L}_\text{rec} = \|\mathbf{I}_\text{comp}-\mathbf{I}_\text{org}\|_{1}$ 
When calculating the final loss for the discriminator, we use a gradient penalty: $\mathcal{L}_\text{reg} = \mathbb{E}_{\mathbf{I}_\text{org}, \mathbf{M}}\Big[\|\nabla \mathcal{D}_\theta(\texttt{stack}(\mathbf{M}, \mathbf{I}_\text{org})\|^2\Big]$. 
The final loss is $\mathcal{L}_{\text{total}} = \mathcal{L}_{\text{adv}} + \lambda_\text{rec}\mathcal{L}_\text{rec}  +\lambda_\text{HRFPL}\mathcal{L}_\text{HRFPL}$. The generator and discriminator are trained adversarially. We empirically set $\lambda_\text{rec}=10$, $\lambda_\text{HRFPL}=5$, and $\lambda_\text{reg}=5$ to balance the order of magnitude of each loss term. 
\section{Experiments}
\label{sec:exp}
\subsection{Datasets and Evaluation Metrics}

We trained seperate models on Places2 and CelebA-HQ datasets. Places2~\cite{places} is a commonly-used dataset containing 8 million training images. We tested our models using the validation set consisting of 36,500 images. CelebA-HQ~\cite{celeba} is a high-quality image dataset of human faces containing 30,000 images. We divided the dataset into a training set with 26,000 images, a validation set with 2,000 images, and a test set with 2,000 images. We followed previous works to use LPIPS~\cite{lpips} and FID~\cite{FID} as the evaluation metrics. We also conducted a user study to evaluate the perceptual quality of our results in a more faithful way. 

\begin{figure*}[t!]
\centering
\includegraphics[width=\linewidth]{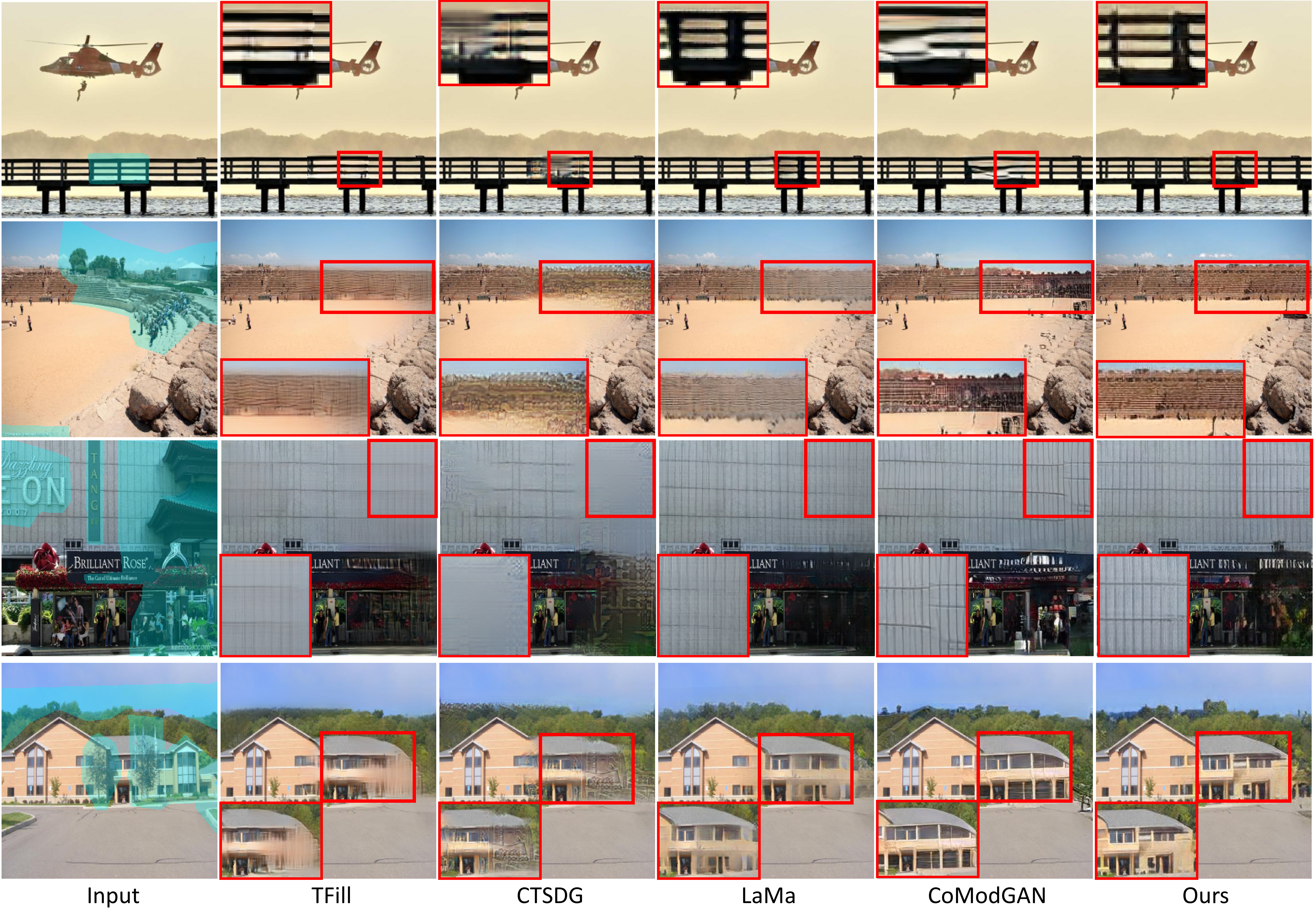}
\vspace{-7mm}
\caption{\textbf{Qualitative comparison to the state-of-the-art methods on Places2:} TFill~\cite{zheng2021tfill}, CTSDG~\cite{iccv_ctsdg}, LaMa~\cite{lama}, CoModGAN$^\dag$~\cite{comodgan}, and our framework (\textit{Ours}). LaMa struggles to generate clear object boundaries while producing fading-out structures. CoModGAN does not have an attention scheme or large receptive field. Thus, it cannot effectively use self-similarity within the image and generates unseen and inconsistent textures. Ours handles structures and textures well in a single model. More results are in the supplementary material. }
\label{fig:places_qual}
\vspace{-0.5cm}
\end{figure*}

\begin{figure*}[t!]
\centering
\includegraphics[width=\linewidth]{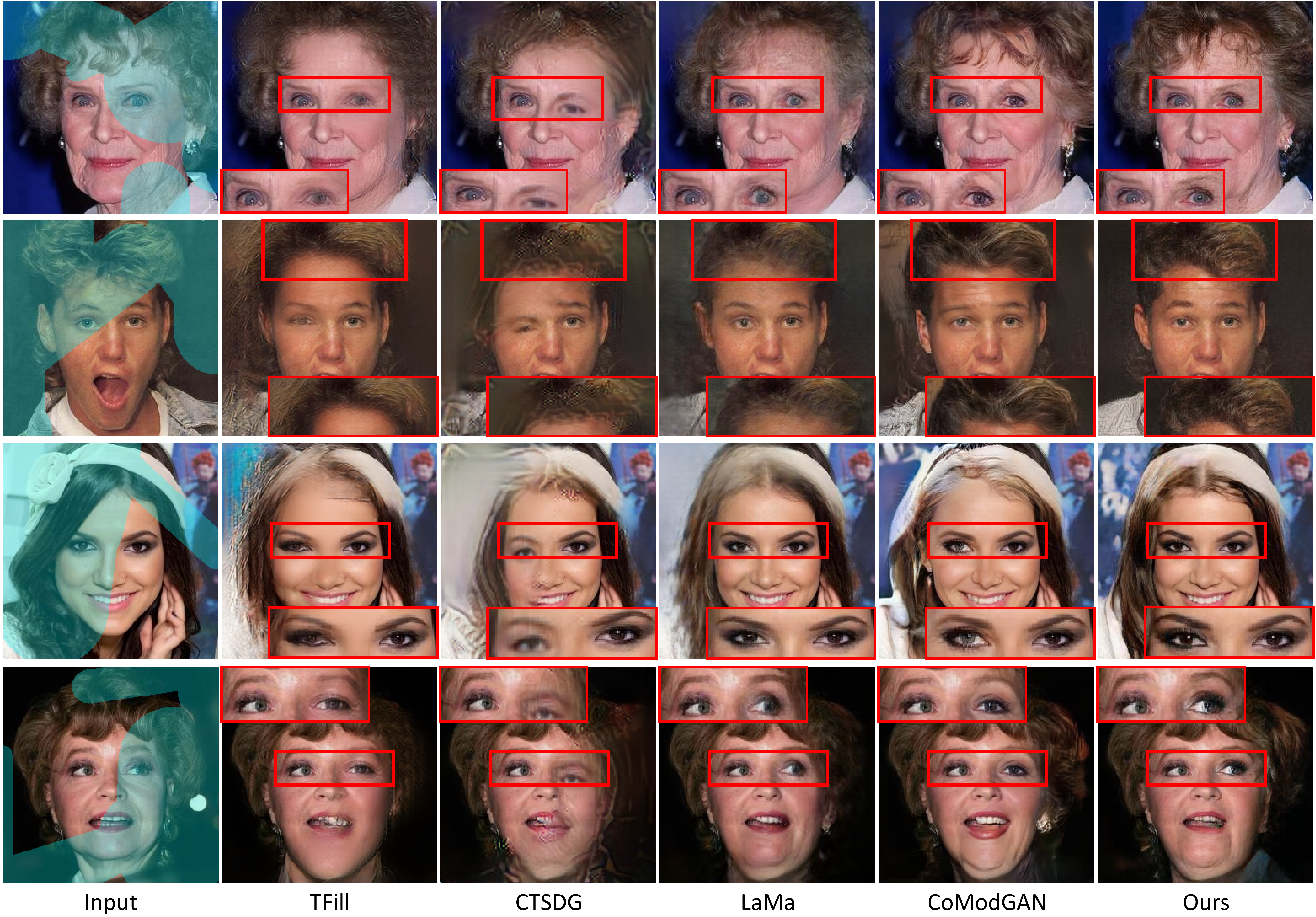}
\vspace{-7mm}
\caption{\textbf{Qualitative comparison to the state-of-the-art methods on CelebA-HQ dataset:} TFill~\cite{zheng2021tfill}, CTSDG~\cite{iccv_ctsdg}, LaMa~\cite{lama}, CoModGAN$^\dag$~\cite{comodgan}, and our framework \textit{Ours}. The images are from the CelebA-HQ val (2k) dataset. LaMa mostly fades out the hair and generates a blurry boundary on the forehead. CoModGAN tends to generate unseen appearances inconsistent with the original face. Zoom in to check the eyes and eyebrows. Ours generate fine-detailed hairs and forehead shapes while preserving the original appearance of the person by generating consistent eyes and gaze direction. More results are in the supplementary material.}
\label{fig:celeba_qual}
\vspace{-0.5cm}
\end{figure*}
\subsection{Implementation Details}

\noindent
\textbf{Network Details.}
The encoder $\mathcal{E}$ downscales the input to a spatial size of $4\!\times\!4$, increasing the channel dimension by $\times2$ at each downscaled resolution to a maximum of $512$ channels. We set the dimension for latent noise vector $\mathbf{z}$ as $512$. We flattened the output from the encoder to a dimension of $1024$ to obtain $\mathbf{z}_\text{enc}$. We set the values of the Number ($L_\text{res}$) of FFC Residual Blocks at different resolutions as $\{L_\text{32}: 1, L_\text{64}: 1, L_\text{128}: 1, L_\text{256}: 1\}$. 
\smallskip

\noindent
\textbf{Training Settings.}
We developed our codebase in PyTorch~\cite{pytorch}. We conducted image completion at $256\!\times\!256$ resolution on the Places2~\cite{places} and CelebA-HQ~\cite{celeba}. We trained our framework and CoModGAN$^\dag$ (for fair comparison) for 25M images on both Places2 and CelebA-HQ. When training on Places2, we randomly cropped $256\!\times\!256$ patches from the high-resolution images during training. We resized the CelebA-HQ images to $256\!\times\!256$, following LaMa~\cite{lama}. We randomly generated free-form masks during training following the generation strategy used in CoModGAN~\cite{comodgan}. We used the Adam~\cite{adam} optimizer with the learning rate set to $0.001$. We used a batch size of $128$.
\smallskip

\noindent
\textbf{Baselines.} We compared our method to various baselines including the milestone works LaMa-Fourier~\cite{lama} and CoModGAN~\cite{comodgan}, a transformer-based work called TFill~\cite{zheng2021tfill}, recent papers addressing structures and textures CTSDG \cite{iccv_ctsdg} and CR-Fill \cite{crfill}, and some older works DeepFill-v2~\cite{deepfillv2} as well as Edge-Connect \cite{edge-connect}, and other well-performed work like AOT-GAN \cite{aotgan}. For most models except for CoModGAN, we used the publicly available codebase and pretrained models. For fair comparison, since the public CoModGAN~\cite{comodgan} checkpoint cannot be tested on $256\!\times\!256$ resolution, we trained our own PyTorch~\cite{pytorch} implementation\footnote{For our CoModGAN$^\dag$ re-implementation, we build on top of the the StyleGAN2~\cite{stylegan2} PyTorch codebase [\href{https://github.com/NVlabs/stylegan2-ada-pytorch\#:\~:text=Training\%20is\%20typically\%205\%25\%E2\%80\%9330,on\%20NVIDIA\%20Tesla\%20V100\%20GPUs.
}{link}] as a more efficient alternative to the old version TensorFlow~\cite{tensorflow} codebase which is validated to be on par with the TensorFlow~\cite{tensorflow} code.} of CoModGAN$^\dag$ with a reconstruction loss and used that for evaluation.

\smallskip

\noindent
\textbf{Evaluation Settings.}
When evaluating on Places2, we resized the images to $256\!\times\!256$ and tested with two different mask strategies: medium and segmentation~\cite{lama} used in LaMa. Basically, the medium masks contain random strokes and rectangle boxes with medium size, and the segmentation masks were computed by replacing the masks of the segmentation onto other positions of the image. Please refer to LaMa~\cite{lama} for more details. We used $30k$ and $4k$ samples for medium and segmentation masks respectively. We evaluated on CelebA-HQ for a total of $2k$ samples with medium and thick mask generation strategy~\cite{lama}.

\subsection{Results and Comparisons}

\noindent
\textbf{Qualitative Results.}
We compared the proposed FcF model to the highly relevant baselines including LaMa \cite{lama}, CoModGAN$^\dag$ \cite{comodgan} (our PyTorch implementation), the latest transformer-based TFill \cite{zheng2021tfill} and the recent structure-texture inpainting network CTSDG \cite{iccv_ctsdg}. The results on Places2 and CelebA-HQ are shown in Fig. \ref{fig:places_qual} and \ref{fig:celeba_qual}. 

As shown in Fig. \ref{fig:places_qual}, our model preserves much better repeating textures compared with CoModGAN. CoModGAN does not have any attention-related modules, so high-frequency features cannot be effectively reused given the limited receptive field. Our model enlarged the receptive field using fast Fourier layers and effectively rendered source textures on newly generated random structures. Meanwhile, ours also outperforms LaMa in generating object boundaries and structures. It is evident that LaMa generates fading-out artifacts when the hole reaches the image or object boundary. LaMa cannot hallucinate good structural information given large holes across longer pixel ranges. Ours, however, leverages the advantages of the coarse-to-fine generator to synthesize a clear shape boundary of objects in a better manner. In conclusion, our model integrates the advantages of two state-of-the-arts and simultaneously generates remarkable structures and textures. 

More qualitative evidence can be found in Fig. \ref{fig:celeba_qual}, which is more intuitive. While testing on face images, especially when we covered half of the faces, LaMa generates fading-out hairs on the forehead, and CoModGAN may use others' eyes to complete the images. Though they both obtain good numbers in the quantitative results, some drawbacks are reflected, making both models not robust enough. Ours demonstrates a sound synthesis of hair and forehead shape and consistent eye and eyebrow appearance like LaMa. We can keep concluding that the proposed model work consistently well on both image structures and consistent textures. 

\begin{table}[t!]
\centering
\setlength{\tabcolsep}{2pt}
    \resizebox{\linewidth}{!}{
    \begin{tabular}{lcccccccc}
\toprule
{}  & \multicolumn{4}{c}{\textbf{Places2} ($\bf 256 \times 256$)} & \multicolumn{4}{c}{\textbf{CelebA-HQ} ($\bf 256 \times 256$)} \\
\cmidrule(lr){2-5}
\cmidrule(lr){6-9}
{} & \multicolumn{2}{c}{\textbf{Medium Masks}} & 
\multicolumn{2}{c}{\textbf{Segm. Masks}} & 
\multicolumn{2}{c}{\textbf{Medium Masks}} & 
\multicolumn{2}{c}{\textbf{Thick Masks}} \\
\cmidrule(lr){2-3}
\cmidrule(lr){4-5}
\cmidrule(lr){6-7}
\cmidrule(lr){8-9}
{\textbf{Method}} & {FID$\downarrow$ } & {LPIPS$\downarrow$ } & {FID$\downarrow$ } & {LPIPS$\downarrow$ } & {FID$\downarrow$ } & {LPIPS$\downarrow$ } & {FID$\downarrow$ } & {LPIPS$\downarrow$ } \\
\midrule
Edge-Connect~\cite{edge-connect} & $3.18$ & $0.131$ & $3.72$ & $0.047$ & $7.15$ & $0.098$ & $8.76$ & $0.122$ \\
DeepFillv2~\cite{deepfillv2} & $3.05$ & $0.129$ & $3.60$ & $0.044$ & $8.10$ & $0.104$ & $9.74$ & $0.119$ \\
AOT-GAN~\cite{aotgan} & $1.95$ & $\bl{0.116}$ & $3.31$ & $0.043$ & $8.27$ & $0.104$ & $13.89$ & $0.135$ \\

CTSDG~\cite{iccv_ctsdg} & $4.58$ & $0.136$ & $4.07$ & $0.047$ & $11.26$ & $0.105$ & $12.38$ & $0.124$ \\
CR-Fill~\cite{crfill} & $3.66$ & $0.129$ & $3.68$ & $0.044$ & --- & --- & --- & --- \\
TFill~\cite{zheng2021tfill} & $2.52$ & $0.120$ & \bl{$3.24$} & $\bl{0.042}$ & $6.49$ & \bl{$0.090$} & $6.54$ & $0.102$ \\
CoModGAN$^\dag$~\cite{comodgan} & \bl{$1.93$} & $0.123$ & $3.41$ & $0.044$ & \bl{$5.86$} & $0.105$ & $\bl{5.82}$ & \bl{$0.091$} \\
LaMa~\cite{lama} & $\mathbf{1.49}$ & $\mathbf{0.109}$ & $\mathbf{2.72}$ & $\mathbf{0.037}$ & \re{$5.18$} & \re{$0.077$} & \re{$5.47$} & $\mathbf{0.080}$ \\
\midrule
FcF (ours) & \re{$1.79$} & \re{$0.114$} & \re{$2.98$} & \re{$0.040$} & $\mathbf{4.42}$ & $\mathbf{0.071}$ & $\mathbf{4.63}$ & \re{$0.086$} \\

\bottomrule
\end{tabular}}
\vspace{-2mm}
\caption{\textbf{Quantitative evaluation on Places2 and CelebA-HQ}. We report LPIPS ($\downarrow$) and  FID ($\downarrow$) metrics. The $\downarrow$ symbol means lower value signifies better performance. The \textbf{bold} text indicates the best performance, followed by \re{red} and \bl{blue} fonts meaning the second and the third place.}
\label{table:sota}
\vspace{-0.3cm}
\end{table}

\begin{table}[ht!]\setlength{\tabcolsep}{3pt}
  \centering
\resizebox{\linewidth}{!}{
\begin{tabular}{lccc}
\toprule
\textbf{Method} & FID$\downarrow$  & LPIPS$\downarrow$ & \begin{tabular}{@{}c@{}} User Preference \\ (Baseline / Equal / Ours) \end{tabular} \\
\midrule
CoModGAN (official)~\cite{comodgan} & $2.32$ & $0.045$ & 21.33\% / 17.33\% / 61.33\%\\
LaMa (official)~\cite{lama} & $2.00$ & $0.040$ & 39.33\% / 12.00 \% / 48.67\%\\
FcFGAN (ours) & $2.06$ & $0.041$ & - / - / - \\
\bottomrule
\end{tabular}}
\vspace{-2mm}
\caption{\textbf{Quantitative Comparisons using $512\!\times\!512$ images on Places2~\cite{places}} for segmentation masks.}
\label{tab:512}
\vspace{-3mm}
\end{table}

\noindent
\textbf{Quantitative Results.} We compared our method to several well-established baselines in \cref{table:sota}. We found that LaMa and ours are always the top-two models and consistently outperform other baseline methods. Other baselines are not proven to work consistently well on larger masks. CoModGAN is not working well on reconstruction. For Places2 evaluation, LaMa is still a strong baseline performing well in FID and the reconstruction-based metric LPIPS. Ours is comparable to the LaMa-Fourier model but is significantly better than the CoModGAN$^\dag$. FFC layers and the proposed FaF-Syn modules add more global features to synthesize repeating textures for better background reconstruction. For the CelebA-HQ dataset, the proposed FcF model sets state-of-the-art while comparing with other baselines. 

Due to the eco-friendly consideration, we include $256\!\times\!256$ resolution synthesis to prove concepts and draw scientific conclusions. In practice, we also trained a model on $512\!\times\!512$ resolution on Places2\cite{places}. We use a batch size of 32 and the $\{L_\text{32}: 1, L_\text{64}: 1, L_\text{128}: 1, L_\text{256}: 1, L_\text{512}: 1\}$ setting while training. We achieve superior performance to the original CoModGAN~\cite{comodgan} and are competitive to LaMa~\cite{lama} as shown in \cref{tab:512}. More qualitative comparison are in the Supplementary material. Thus, we demonstrate that our framework generalizes to higher resolutions equally well.

\noindent
\textbf{User Study.} The existing metric LPIPS is hard to capture the enhanced textures and variant structures given complex scenes in Places2 \cite{places}. FID is neither an ideal metric when we achieve equally good performance as LaMa~\cite{lama} on man-made scenes in Places2~\cite{places}. To further validate our model advantages, we conduct a user study via Amazon Mechanical Turk with 150 real user cases at $512\times 512$ resolution. We let the users choose 'better', 'equal', or 'worse'. As shown in \cref{tab:512}, 
our preference rate is the best, which further demonstrates our better visual quality.

\subsection{Ablation Studies}
\label{subsec:ablat}

\begin{table}[t!]\setlength{\tabcolsep}{10pt}
  \centering
  \footnotesize
  \label{table:ablation_ffc_res}
  \resizebox{\linewidth}{!}{
  \begin{tabular}{ccccccc}
\toprule
\cmidrule(lr){1-4}
$\mathbf{L_{32}}$ & $\mathbf{L_{64}}$ & $\mathbf{L_{128}}$ & $\mathbf{L_{256}}$  & FID $\downarrow$ & LPIPS $\downarrow$ \\
\midrule

0 & 0 & 0 & 0 & 13.53 & 0.275 \\
0 & 1 & 1 & 1 & 12.14 & 0.266 \\
0 & 1 & 2 & 2 & 11.92 & \textbf{0.263} \\
0 & 2 & 2 & 2 & 12.77 & 0.268 \\
1 & 1 & 1 & 1 & \textbf{11.33} & 0.264 \\
2 & 2 & 2 & 2 & 15.33 & 0.280 \\
\bottomrule
\end{tabular}

}
  \vspace{-0.2cm}
  \caption{\textbf{Ablation on number of FFC Residual Blocks.} We find $\{L_\text{32}: 1, L_\text{64}: 1, L_\text{128}: 1, L_\text{256}: 1\}$ performs the best on the FID and LPIPS metrics.
  }
    \vspace{-0.4cm}
\end{table}

\begin{table}[t!]\setlength{\tabcolsep}{25pt}
  \centering
  \footnotesize
  \label{table:ablation_struct}
  \resizebox{\linewidth}{!}{
  \begin{tabular}{lcc}
\toprule
\footnotesize
\textbf{Module} & FID $\downarrow$ & LPIPS $\downarrow$ \\
\midrule
FaF-Syn (ours) & \textbf{11.33} & 0.264 \\
FFC with $\mathbf{X}_\text{skip}$ & 11.97 & 0.267 \\
FaF-Res with $\mathbf{X}_\text{skip}$ & 12.58 & 0.267 \\
w.o. FFC & 13.53 & 0.275 \\

\bottomrule
\end{tabular}}
  \vspace{-0.2cm}
  \caption{\textbf{Ablation on Structures.} Our FaF-Syn performs the best on the FID and LPIPS metrics. The results show the effectiveness of the proposed design.
  }
  \label{tab:fafsyn}
\vspace{-0.6cm}
\end{table}

\begin{figure}[t!]
\centering
\includegraphics[width=\linewidth]{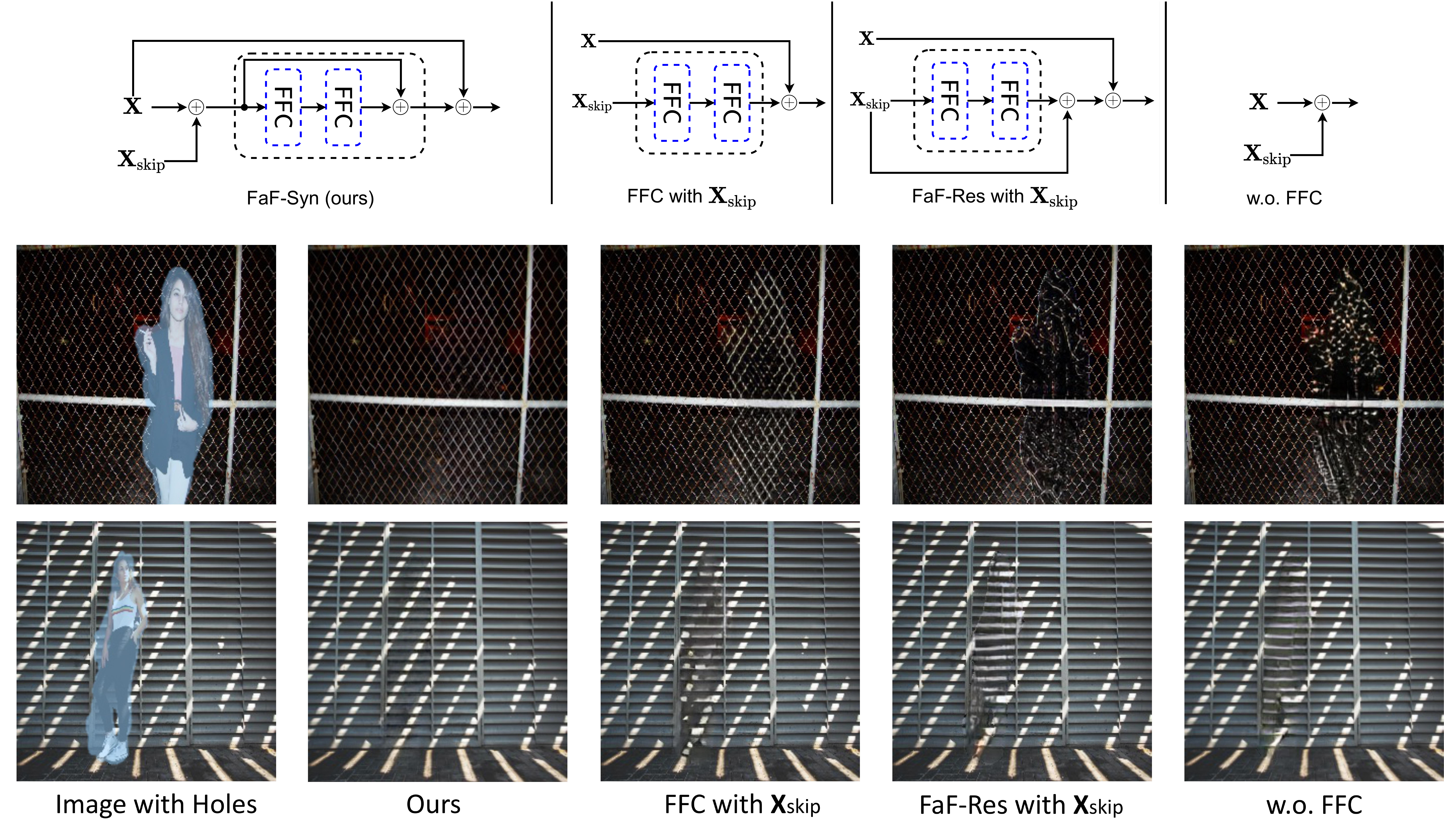}
\vspace{-0.8cm}
\caption{\textbf{Ablation study on alternatives of FaF-Syn module.} The results show the necessity of merging
X before feeding $\mathbf{X}_\text{skip}$ into the FaF-Syn Residual block.} 
\label{fig:abla}
\vspace{-0.3cm}
\end{figure}

\noindent
\textbf{Ablation on Number of FFC Residual Blocks.} The number of FFC Residual Blocks inside our FaF-Res Block is an important tunable hyper-parameter. We experiment with various settings for $\{L_\text{32}, L_\text{64}, L_\text{128}, L_\text{256}\}$ in \cref{table:ablation_ffc_res}. We empirically observe that the setting $\{L_\text{32}: 1, L_\text{64}: 1, L_\text{128}: 1, L_\text{256}: 1\}$ gives the best performance.

\noindent
\textbf{Ablation on FaF-Syn Structures.}
We illustrated the inpainting results generated by different options of FaF-Syn module connections. We merged the encoder and decoder features in our current design before feeding them into the FaF-Syn Residual block. Alternatively, we experimented with two different ways: (1) directly connecting the FFC layers with the skipped features $\mathbf{X}_{skip}$ from the encoder (similar to using FFC inside encoder), or (2) connecting the skipped features with the FaF-Syn Residual block before merging to the generator feature $\mathbf{X}$. The qualitative results in \cref{fig:abla} and quantitative comparison in \cref{tab:fafsyn}  show the necessity of merging $\mathbf{X}$ and $\mathbf{X}_{skip}$ before feeding it into the FaF-Syn Residual block.

\section{Conclusion}
\vspace{-0.5em}

This work tackles the persistent challenges of synthesizing fair structures and textures in the hole regions. To this end, we propose a Fourier Coarse-to-Fine (FcF) inpainting framework that unites the receptive power of fast fourier convolutions to capture global repeating textures with the co-modulated coarse-to-fine generator to generate realistic image structures. Specifically, we proposed a simple yet effective FaF-Syn module aggregating the features from both the encoder and the generator to render textures on the generated structures progressively. Our model achieved a new state-of-the-art performance on the CelebA-HQ dataset and the best perceptual quality on the Places2 dataset. Extensive qualitative and quantitative analysis indicated that our framework is relatively robust to large masks and does not generate fading-out artifacts.

{\small
\bibliographystyle{ieee_fullname}
\bibliography{main}
}

\appendix
\begin{center}{\bf \Large Appendix}\end{center}
\renewcommand{\thetable}{\Roman{table}}
\renewcommand{\thefigure}{\Roman{figure}}
\setcounter{table}{0}
\setcounter{figure}{0}

\Crefname{appendix}{Appendix}{Appendixes}

\noindent
We provide ablation studies on applying the FaF-Syn module to different resolutions and loss functions in \cref{sec:supp_ablat}. We also provide a quantitative comparison at different masked ratios in \cref{sec:mask_ratio}. Lastly, we provide more qualitative comparisons in \cref{sec:qual}.

\section{Additional Ablation Studies}
\label{sec:supp_ablat}

\noindent
\textbf{Ablation on Resolution for FFC Residual Blocks.} We experiment with application of our FaF-Syn block at lower resolutions with the setting $\{L_\text{32}: 1,, L_\text{64}: 1, L_\text{128}: 1, L_\text{256}: 1\}$. For each experiment we set $L_\text{res} = 1$ for res $\in \{8,16\}$. We observe that adding FFC to lower resolutions harms the performance as shown in \cref{tab:ablation_res}. We reason that the lower resolution features contain insufficient spatial information required for modeling the global context. The coarse-level features input to the FFC are magnified with noise, thus leading to a drop in performance and even instability during training $(4\times4)$.

\begin{table}[ht!]\setlength{\tabcolsep}{10pt}
  \centering
  \resizebox{1.0\linewidth}{!}{
  \begin{tabular}{cccccccc}
\toprule
$8\!\times\!8$ & $16\!\times\!16$ & $32\!\times\!32$ & $64\!\times\!64$ & $128\!\times\!128$ & $256\!\times\!256$ & \textbf{FID} & \textbf{LPIPS} \\
\midrule
 & & & $\checkmark$ & $\checkmark$ & $\checkmark$ & 12.14 & 0.266 \\
 & & $\checkmark$ & $\checkmark$ & $\checkmark$ & $\checkmark$ & \textbf{11.33} & 0.264 \\
  & $\checkmark$ & $\checkmark$ & $\checkmark$ & $\checkmark$ & $\checkmark$ & 11.69 & 0.263  \\
 $\checkmark$ & $\checkmark$ & $\checkmark$ & $\checkmark$ & $\checkmark$ & $\checkmark$ & 12.24 & 0.269  \\
$\checkmark$ & & & $\checkmark$ & $\checkmark$ & $\checkmark$ & 11.83 & 0.266 \\
& $\checkmark$ & & $\checkmark$ & $\checkmark$ & $\checkmark$ & 11.44 & 0.262 \\

\bottomrule
\end{tabular}}
  \caption{\textbf{Ablation on resolution for FFC Residual Blocks.} Applying FFC to lower resolution coarse-features harms the performance.
  }
  \label{tab:ablation_res}
\end{table}

\begin{table}[t!]\setlength{\tabcolsep}{10pt}
  \centering
  \resizebox{0.7\linewidth}{!}{
  \begin{tabular}{cccc}
\toprule
$\mathcal{L}_\text{rec}$ & $\mathcal{L}_\text{HRFPL}$ & FID $\downarrow$ & LPIPS $\downarrow$ \\
\midrule

 & & $16.83$ & $0.297$  \\
$\checkmark$ & & $14.14$ & $0.279$ \\
 & $\checkmark$ & $12.52$ & $0.270$ \\
$\checkmark$ & $\checkmark$ & $\mathbf{11.33}$ & $\mathbf{0.264}$ \\
\bottomrule
\end{tabular}}
  \caption{\textbf{Ablation on Loss Functions.} We study the impact of reconstruction and HRFPL losses during training. We observe that pixel and feature level supervision is critical to the success of FFC based networks.
  }
  \label{tab:ablation_loss}
\end{table}

\noindent
\textbf{Loss Functions.} We ablate the effect of different loss terms on the inpainting performance of our framework. We remove the $\mathcal{L}_\text{rec}$ and $\mathcal{L}_\text{HRFPL}$ from the total loss to study the importance of pixel and feature level supervision, respectively. We use the $L_\text{adv}$ and the $R_1$ regularization as usual. We trained our models for 10M images and evaluated them on 10k images with free-form masks ~\cite{comodgan} sampled from the Places2~\cite{places} val dataset. We observe an increase in the FID and LPIPS scores when removing the loss terms. The major drop in performance (increase in FID and LPIPS score) happens when we remove both the loss terms. We also conclude that using only adversarial loss while training models based on FFC~\cite{ffc} can lead to major drop as FFC requires supervision from the frequency signal present in the images as shown in \cref{tab:ablation_loss}.

\begin{figure*}[t!]
\begin{subfigure}{0.49\linewidth}
  \centering
  \includegraphics[width=\linewidth]{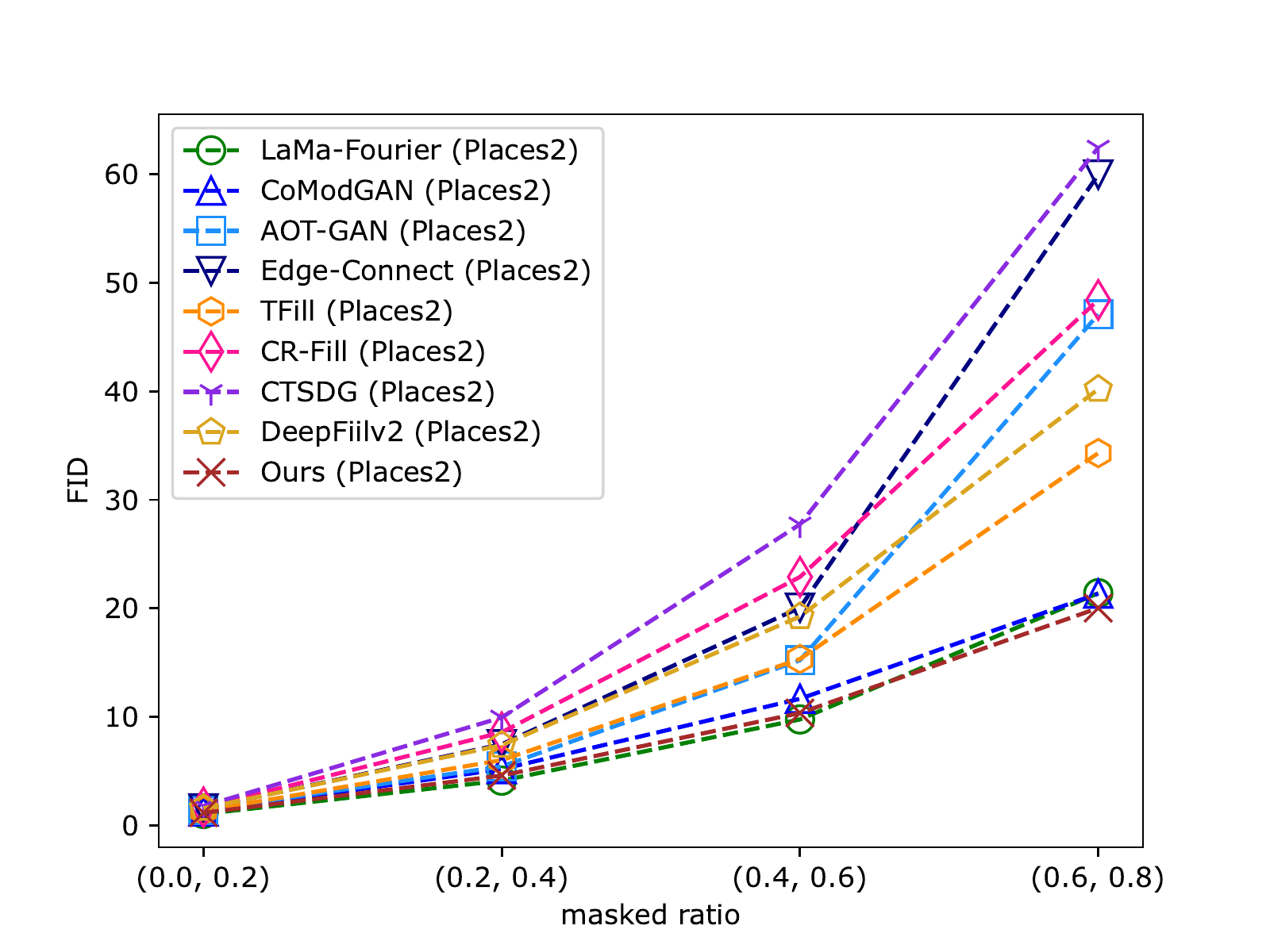}
  \subcaption{FID comparison on Places2}
  \label{fig:places_fid_graph}
\end{subfigure}
\begin{subfigure}{0.49\linewidth}
  \centering
  \includegraphics[width=\linewidth]{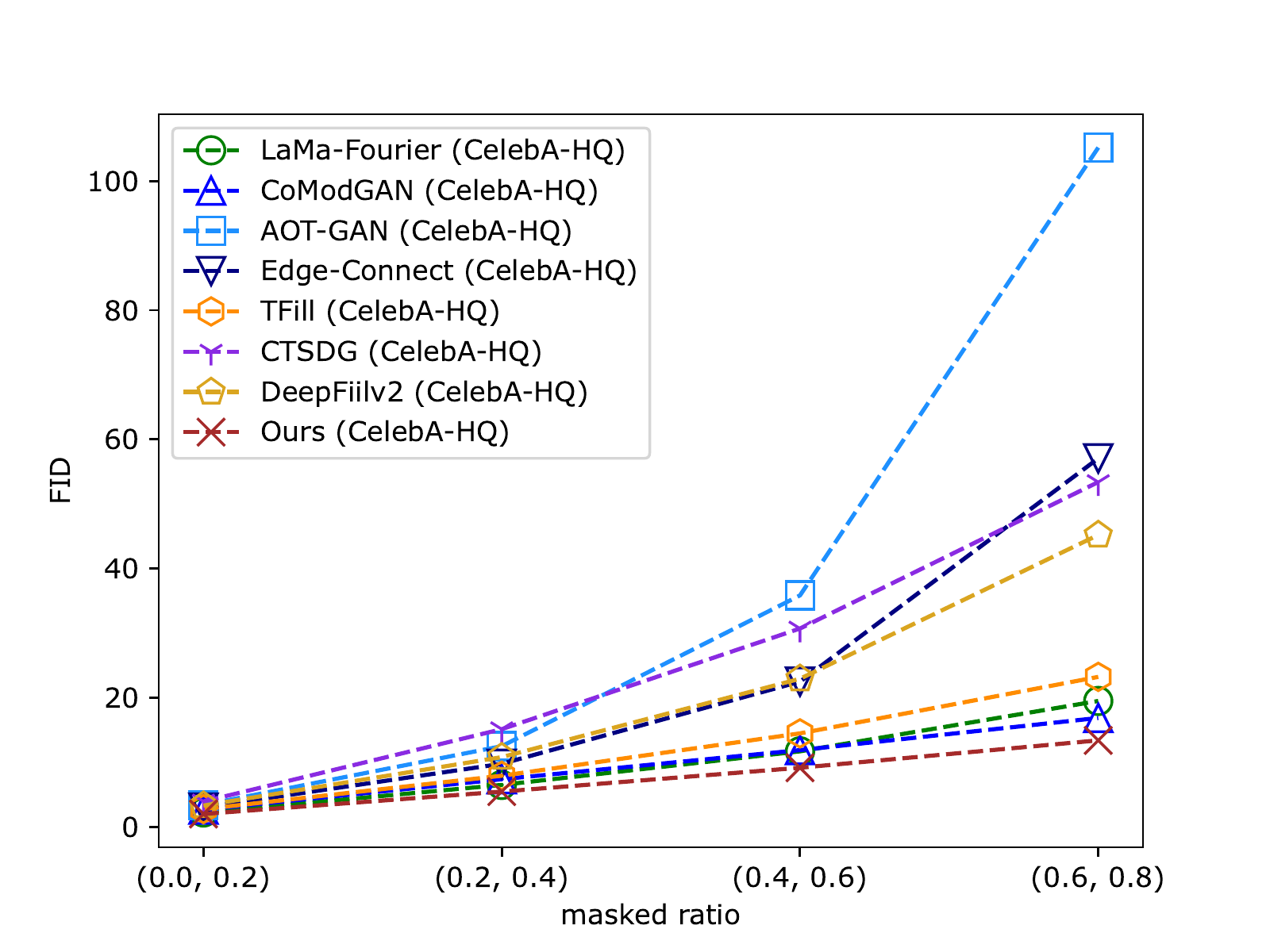}
  \subcaption{FID Comparison on CelebA-HQ}
  \label{fig:celeb_fid_graph}
\end{subfigure}
\begin{subfigure}{0.49\linewidth}
  \centering
  \includegraphics[width=\linewidth]{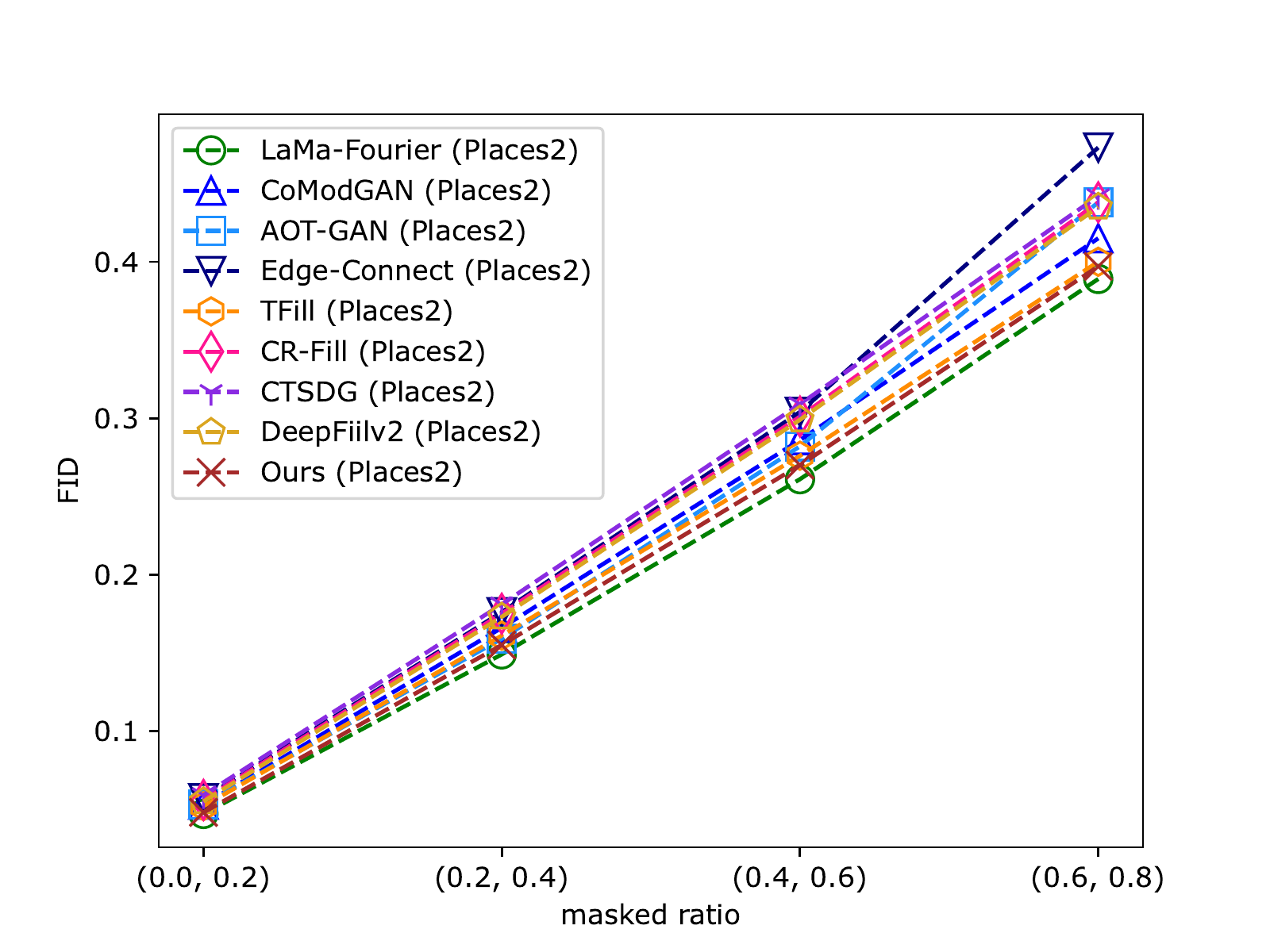}
  \subcaption{LPIPS comparison on Places2}
  \label{fig:places_lpips_graph}
\end{subfigure}
\begin{subfigure}{0.49\linewidth}
  \centering
  \includegraphics[width=\linewidth]{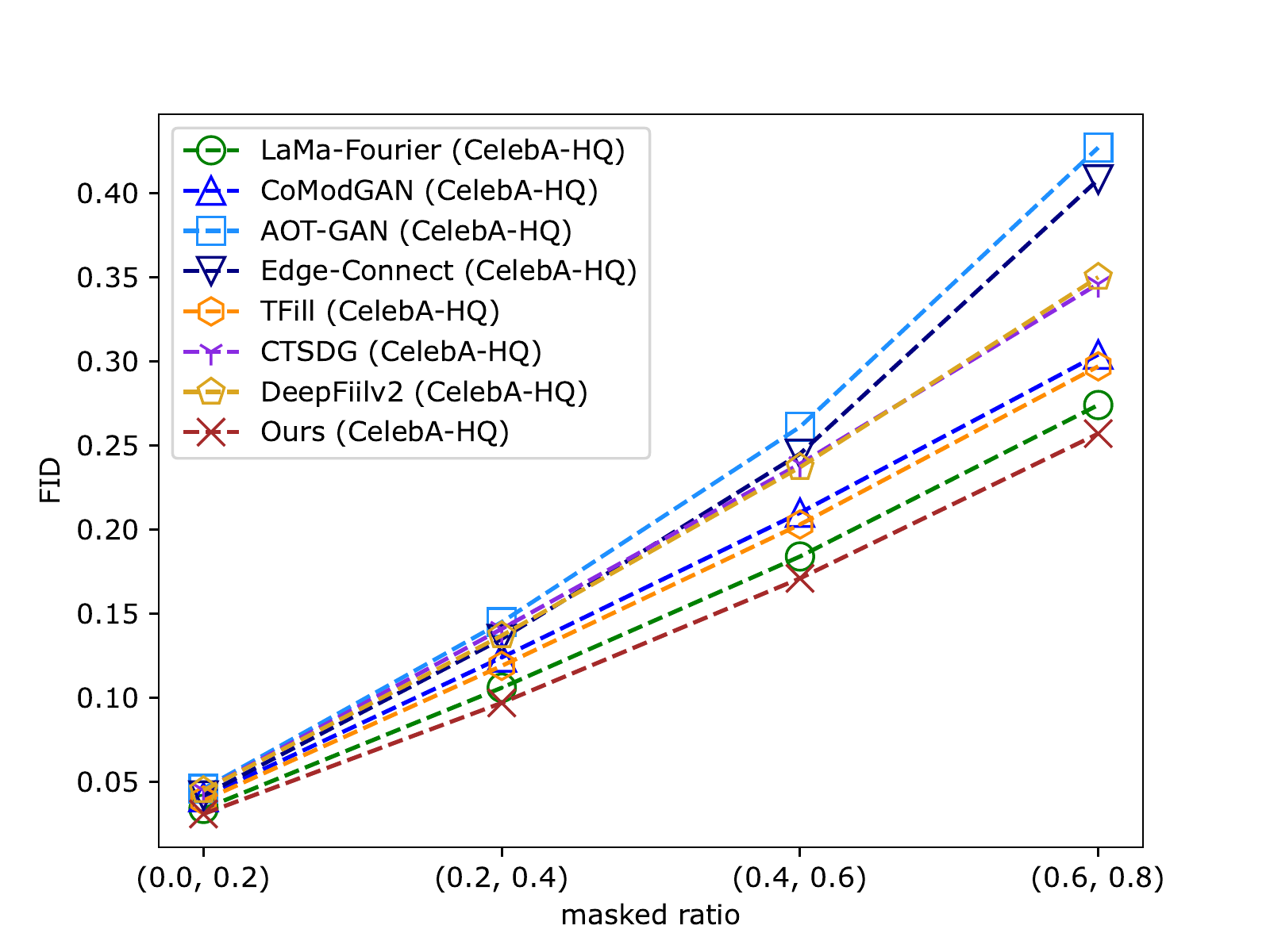}  
  \subcaption{LPIPS comparison on CelebA-HQ}
  \label{fig:celeba_lpips_graph}
\end{subfigure}
\caption{\textbf{Evaluation on ratio-wise masks.} We plot and compare the FID and LPIPS scores of our framework to all baselines with respect to masked ratios. Larger masks bring more challenging cases in completing structures. Ours, as well as LaMa and CoModGAN$^\dag$, perform consistently well than other baselines. }
\label{fig:graphs}
\end{figure*}

\section{Quantitative comparison at different Masked Ratios.} 
\label{sec:mask_ratio}

We study the quantitative performance with different hole ratios in \cref{fig:graphs}. A larger hole means it is more challenging to complete the structure. We use a free-form mask generation strategy to generate 10k samples for Places2 and 2k samples for CelebA-HQ during evaluation. The results showed that only Ours, LaMa~\cite{lama}, and CoModGAN$^\dag$~\cite{comodgan} performed consistently well as the hole size increased. Other state-of-the-arts still struggle to fill complex structures. Among them, TFill~\cite{zheng2021tfill} with transformer-based network structures works better. Ours are robust enough for both Places2~\cite{places} and CelebA-HQ~\cite{celeba} datasets.
\section{More Qualitative Results}
\label{sec:qual}

\begin{figure*}[ht!]
\centering
\includegraphics[width=0.97\linewidth]{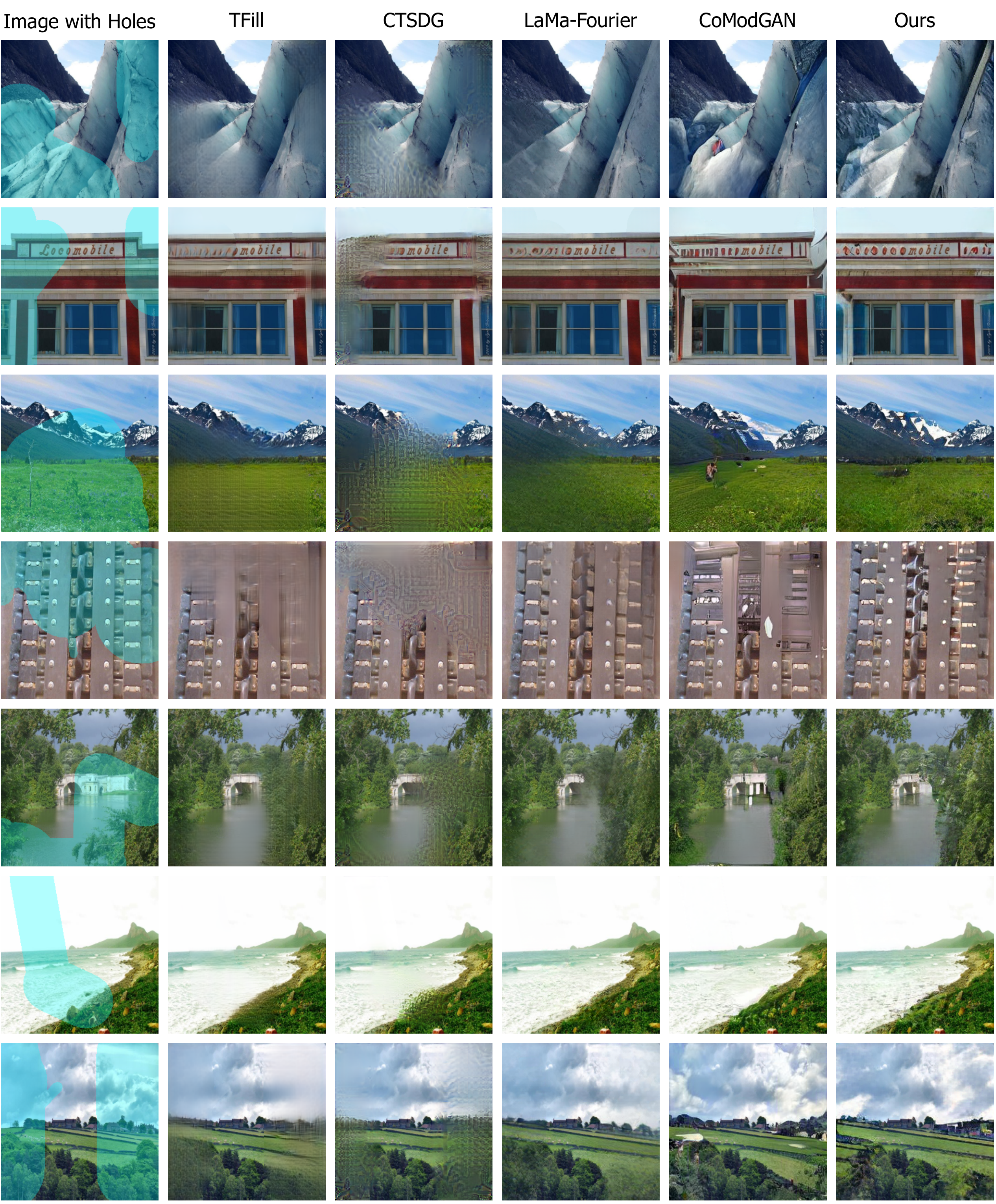}
\caption{\textbf{Qualitative examples for image completion on $256\!\times\!256$ Places2.} We compare texture and structure completion among TFill~\cite{zheng2021tfill}, CTSDG~\cite{iccv_ctsdg}, LaMa~\cite{lama}, CoModGAN$^\dag$~\cite{comodgan}, and FcF (\textit{Ours})}
\label{fig:supp_places_qual}
\end{figure*}

\begin{figure*}[ht!]
\centering
\includegraphics[width=1.0\linewidth]{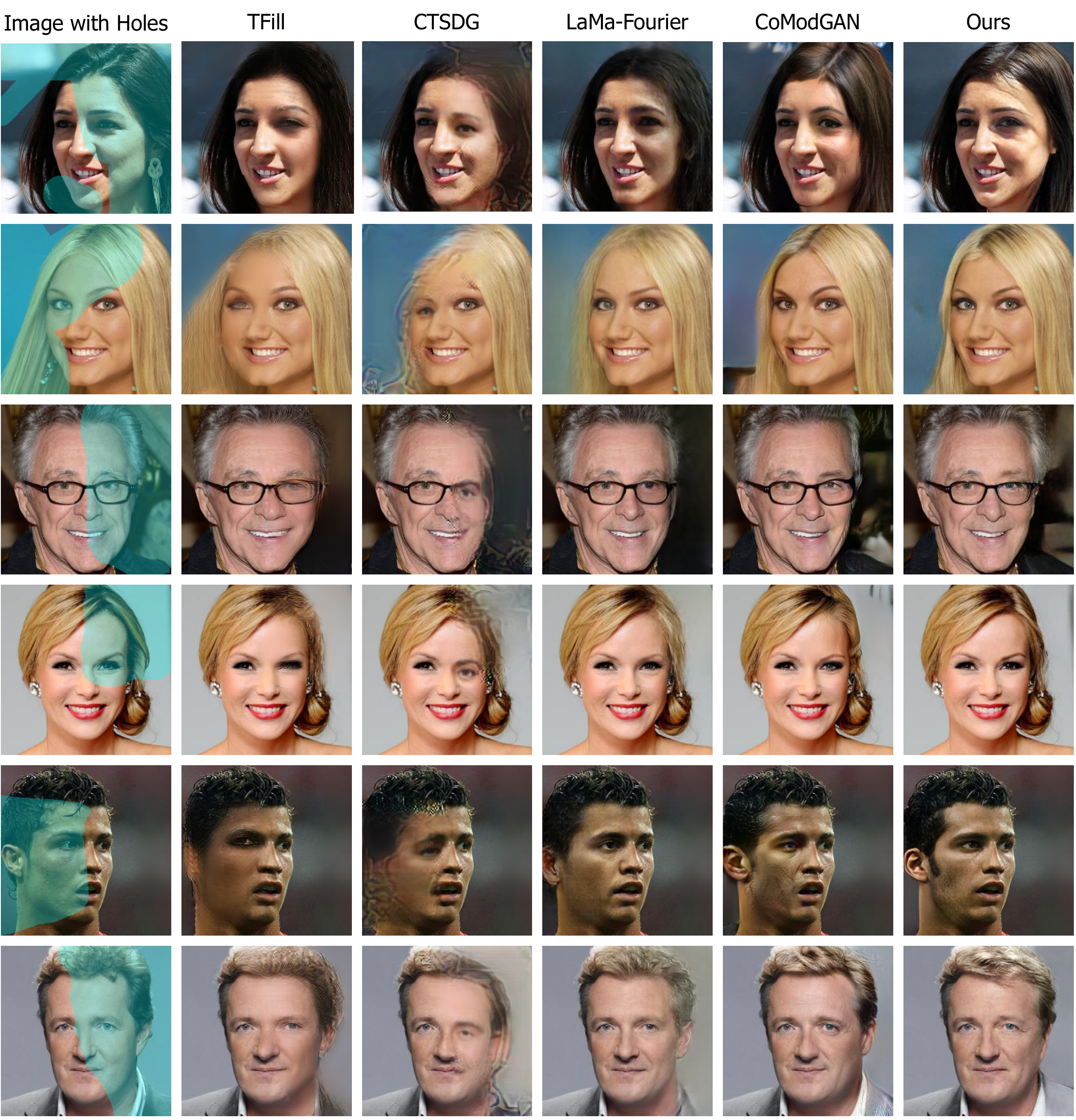}
\caption{\textbf{Qualitative examples for image completion on $256\!\times\!256$ CelebA-HQ.} We compare the face structure completion among TFill~\cite{zheng2021tfill}, CTSDG~\cite{iccv_ctsdg}, LaMa~\cite{lama}, CoModGAN$^\dag$~\cite{comodgan}, and FcF (\textit{Ours})}
\label{fig:supp_celeba_qual}
\end{figure*}

We provide more qualitative results on Places2~\cite{places} and CelebA-HQ~\cite{celeba} in \cref{fig:supp_places_qual} and \cref{fig:supp_celeba_qual}, respectively. We compare our FcF framework to TFill~\cite{zheng2021tfill}, CTSDG~\cite{iccv_ctsdg}, LaMa-Fourier~\cite{lama} and CoModGAN$^\dag$ (our PyTorch~\cite{pytorch} implementation). 





\begin{figure*}[ht!]
\centering
\includegraphics[width=0.85\linewidth]{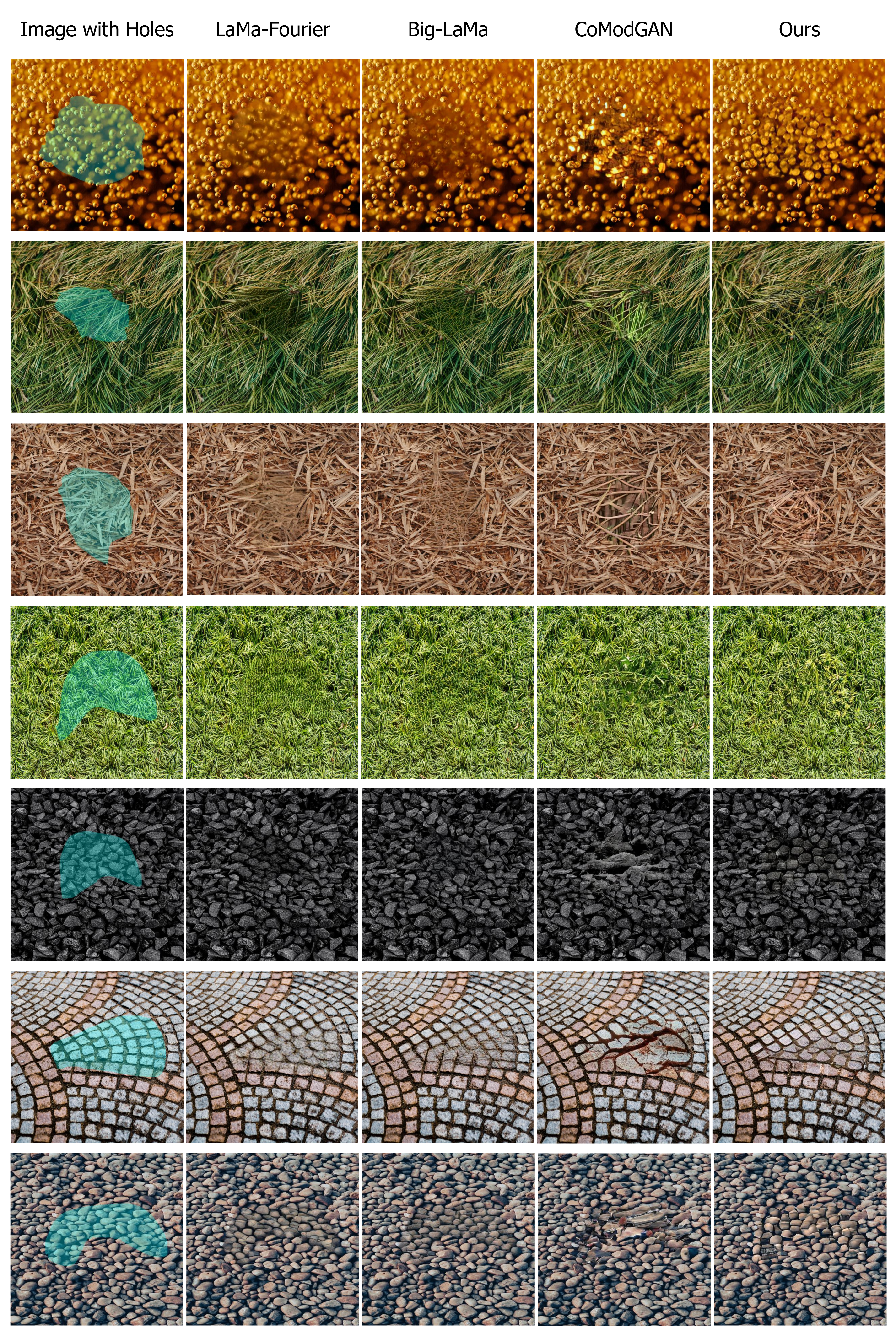}
\vspace{-10pt}
\caption{\textbf{Qualitative examples for image completion on $512\!\times\!512$ Texture Images.} We compare texture and structure completion among LaMa-Fourier~\cite{lama}, Big-LaMa~\cite{lama}, CoModGAN$^\dag$~\cite{comodgan}, and FcF (\textit{Ours}). Zoom-in for best view.}
\label{fig:supp_text_qual}
\end{figure*}
\begin{figure*}[ht!]
\centering
\includegraphics[width=0.85\linewidth]{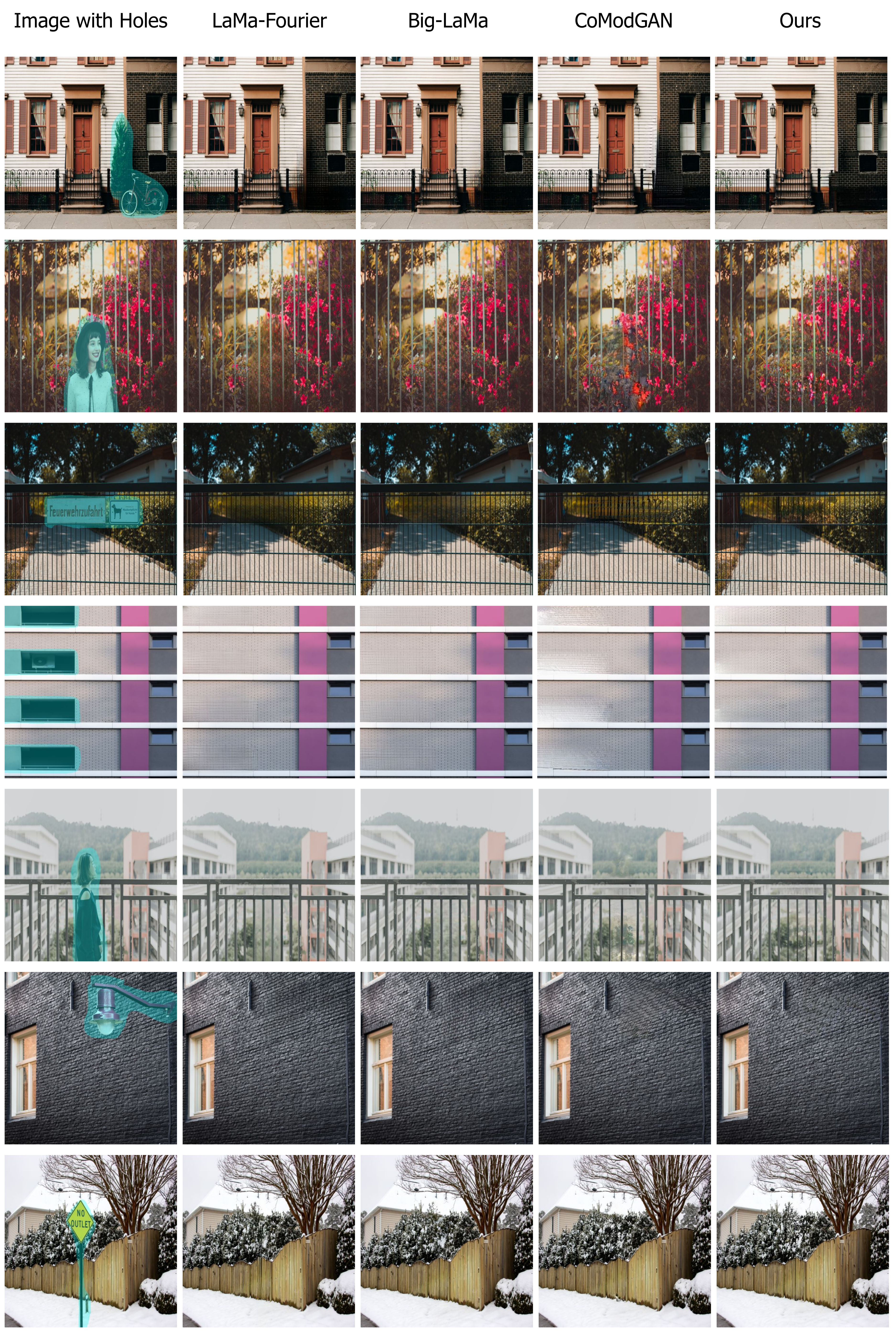}
\vspace{-10pt}
\caption{\textbf{Qualitative examples for image completion on $512\!\times\!512$ images.} We compare texture and structure completion among LaMa-Fourier~\cite{lama}, Big-LaMa~\cite{lama}, CoModGAN$^\dag$~\cite{comodgan}, and FcF (\textit{Ours}). Zoom-in for best view.}
\label{fig:supp_obj_qual}
\end{figure*}
\begin{figure*}[ht!]
\centering
\includegraphics[width=1.0\linewidth]{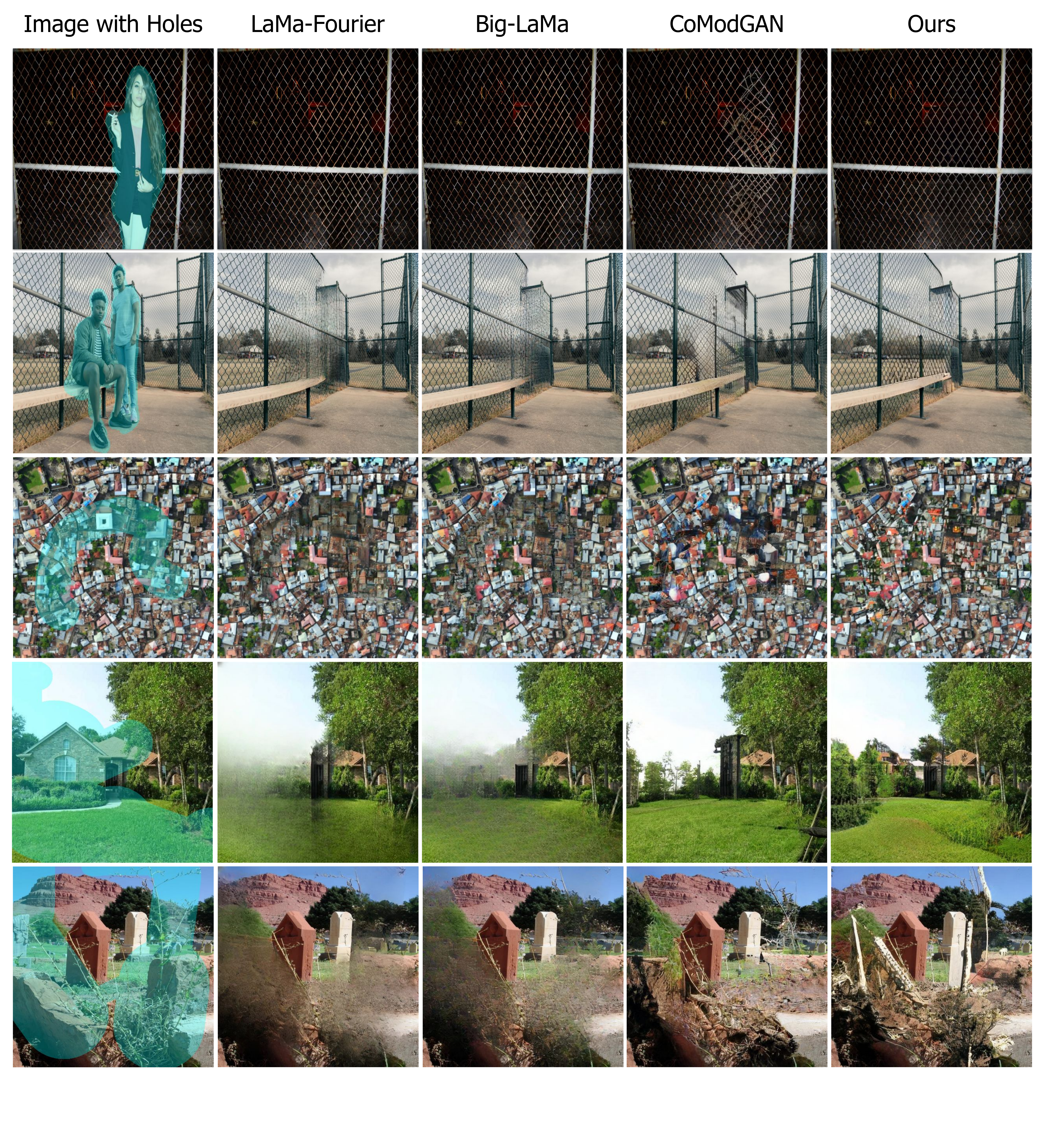}
\vspace{-50pt}
\caption{\textbf{Qualitative examples for image completion on $512\!\times\!512$ images.} We compare texture and structure completion among LaMa-Fourier~\cite{lama}, Big-LaMa~\cite{lama}, CoModGAN$^\dag$~\cite{comodgan}, and FcF (\textit{Ours}). Zoom-in for best view.}
\label{fig:supp_places_512_qual}
\end{figure*}

We also provide qualitative comparisons for our model trained on $512\!\times\!512$ resolution to the official publicly released models: LaMa-Fourier~\cite{lama}, Big-LaMa~\cite{lama} and CoModGAN~\cite{comodgan} in \cref{fig:supp_places_512_qual}, \cref{fig:supp_text_qual}, and
 \cref{fig:supp_obj_qual}.

\end{document}